# The Evolution of Alpha in Finance Harnessing Human Insight and LLM Agents

Mohammad Rubyet Islam[0000-0002-8263-7560]

George Mason University, Fairfax VA 22030, USA
mislam30@gmu.edu

**Abstract -** The pursuit of alpha—returns that exceed market benchmarks—has undergone a profound transformation, evolving from intuition-driven investing to autonomous, AI-powered systems. This paper introduces a comprehensive five-stage taxonomy that traces this progression across manual strategies, statistical models, classical machine learning, deep learning, and agentic architectures powered by large language models (LLMs). Unlike prior surveys focused narrowly on modeling techniques, this review adopts a system-level lens, integrating advances in representation learning, multimodal data fusion, and tool-augmented LLM agents. The strategic shift from static predictors to context-aware financial agents capable of real-time reasoning, scenario simulation, and cross-modal decision-making is emphasized. Key challenges in interpretability, data fragility, governance, and regulatory compliance—areas critical to production deployment—are examined. The proposed taxonomy offers a unified framework for evaluating maturity, aligning infrastructure, and guiding the responsible development of next-generation alpha systems.

## 1 Introduction

### 1.1 The Evolving Pursuit of Alpha

Achieving alpha—returns that outperform risk-adjusted market benchmarks—has traditionally stood at the heart of investment strategy and portfolio management [1]. Traditionally, alpha generation relied on human expertise, intuition, and narrative reasoning. Portfolio managers like Benjamin Graham and Warren Buffett built strategies rooted in fundamentals [2,3]. Technical analysis later emerged, emphasizing price trends and momentum signals as proxies for investor behavior [4,5]. These early methods were powerful but lacked scalability, objectivity, and testability. As markets became more complex and data-intensive, manual approaches gave way to statistical models grounded in asset pricing theory [1, 6-8] and later to algorithmic trading systems [9,10]. A formal representation of alpha, widely known as Jensen's Alpha, refines the Capital Asset Pricing Model (CAPM) to isolate abnormal returns (Equation 1):

$$\alpha = R_i - [R_f + \beta_i(R_m - R_f)] \tag{1}$$

where $R_i$ is the return of the investment, $R_f$ the risk-free rate, $R_m$ the market return, and $\beta_i$ the sensitivity of the investment to market movements.

This equation, first introduced by Michael Jensen, quantifies the excess return not explained by systematic market exposure [6]. The rise of hedge funds and the explosion of alternative data catalyzed further advances in systematic strategies [11,12]. Most recently, machine learning has enabled predictive models across asset classes and risk domains [13-15]. Today, alpha generation is undergoing another paradigm shift—toward agentic AI powered by large language models (LLMs) capable of reasoning, tool use, and multimodal data integration in real time [16-20].

## 1.2 Why This Review Is Timely

Recent advances in LLMs—such as GPT-4, LLaMA 2, FinGPT, and BloombergGPT—have significantly expanded the boundaries of automated finance. These models now enable sentiment-aware forecasting, document summarization, event simulation, and interactive reasoning over structured and unstructured data [18-22]. More importantly, the rise of **agentic architectures,** such as those supported by LangChain and AutoGPT, has transformed LLMs from passive predictors into autonomous systems capable of executing complex, multi-step tasks [23,24]. Despite these advancements, the literature on alpha generation remains fragmented. Prior surveys focus on machine learning, deep learning, or isolated NLP applications, without a cohesive view of how LLMs integrate into the full alpha pipeline [25-27]. Furthermore, few studies address the regulatory, ethical, and operational implications of deploying autonomous agents in high-stakes financial environments—gaps that are increasingly relevant as institutions like JPMorgan and Bloomberg integrate LLMs into production systems, and regulators such as the SEC and ESMA issue new AI guidelines [28,29]. A recent treatment of these ethical and governance concerns, especially in the context of cyber-physical risk and financial autonomy, is offered in *Generative AI, Cybersecurity, and Ethics* by Rubyet Islam [30].

## 1.3 Contributions

This review seeks to fill important gaps in the existing literature—particularly those not addressed by earlier surveys such as Cao [25] —by presenting a comprehensive framework that captures the historical, technical, and regulatory evolution of alpha generation. In contrast to Cao's model-centric focus on machine learning pipelines, this review adopts a systems-level perspective, emphasizing the progression from manual strategies to agentic architectures driven by large language models. The main contributions are as follows:

- A structured overview of alpha strategy evolution, formalized through a five-stage taxonomy spanning manual heuristics to autonomous LLM-driven agents.

- A structured framework for assessing alpha generation systems based on their level of automation, modeling complexity, and decision-making intelligence.

- An integrated review of quantitative finance, NLP, and AI governance developments, highlighting key deployment considerations in institutional finance.

### 1.4 Structure of this Paper

The remainder of the paper is structured as follows. Section 2 defines key concepts and highlights the importance of alpha in quantitative finance, while tracing its evolution from manual strategies to statistical modeling and classical machine learning. Section 3 delves into deep learning and multimodal approaches. Section 4 examines the role of large language models (LLMs), both as predictive tools and autonomous agents. Section 5 presents our proposed maturity taxonomy. Section 6 addresses cross-cutting challenges, and Section 7 outlines directions for future research.

## 2 Foundations and Early Evolution

### 2.1 Manual and Statistical Alpha

The earliest approaches to alpha generation were grounded in human intuition, qualitative reasoning, and discretionary decision-making. Investment legends exemplified this stage, using valuation metrics, company fundamentals, and macroeconomic trends to identify long-term opportunities [2, 3, 31]. Technical analysis emerged alongside, emphasizing price patterns and behavioral indicators like RSI (Relative Strength Index) and MACD (Moving Average Convergence Divergence) [4,5]. Table 1 summarizes the comparative features of manual, technical, and statistical alpha strategies and links directly to the discussion in Sections 5 (Stage 1 and Stage 2).

**Table 1.** Manual, Technical, and Statistical Alpha Strategies

| Strategy type | Key techniques | Data used | Strengths | Limitations |
|---|---|---|---|---|
| Manual / Fundamental | Valuation, narrative reasoning, margin of safety, site visits | Balance sheets, reports, macro data | Deep insight, long-term conviction | Non-scalable, subjective, not easily testable |
| Technical Analysis | RSI, MACD, moving averages, price patterns | Historical prices, volumes | Fast execution, behavioral insights | Lacks theoretical rigor, vulnerable to false signals |
| Statistical Models | CAPM, APT, Fama-French, regression, factor modeling | Structured financial and macro data | Interpretable, scalable, risk decomposition | Linearity assumptions, model fragility under shifts |

As markets became more data-intensive, statistical models such as Capital Asset Pricing Model (CAPM), Arbitrage Pricing Theory (APT), and the Fama-French factor framework [26] brought structure and scalability to alpha strategies [1,7, 8] These models allowed systematic risk attribution and return decomposition but introduced their own limitations— chiefly their reliance on linear relationships, sensitivity to model specification, and dependence on structured financial data. As a result, they often failed to capture non-linear effects, regime shifts, and unstructured signals increasingly relevant in modern markets.

## 2.2 Classical Machine Learning

To overcome the rigidity of traditional statistical models, the field advanced toward classical machine learning (ML)—a data-driven paradigm capable of uncovering non-linear relationships and hidden structures in high-dimensional financial data. Algorithms such as Random Forests, XGBoost, and Support Vector Machines (SVMs) gained prominence for tasks like return prediction, factor mining, anomaly detection, and earnings surprise classification [32,34, 35]. Unsupervised methods including k-means clustering and k-nearest neighbors (k-NN) proved useful in regime detection, market segmentation, and peer group analysis without predefined labels [15]. Table 2 summarizes core classical ML algorithms frequently applied in alpha generation pipelines, outlining their methodological strengths, practical use cases, and key limitations.

**Table 2.** Classical ML Algorithms for Alpha Generation [20, 27,34-38]

| Algorithm | Type | Key Strengths | Common Use in Finance | Limitation |
|---|---|---|---|---|
| Random Forest | Ensemble (Trees) | Robust to noise, ranks feature importance | Factor mining, return prediction | Low interpretability |
| XGBoost | Boosted Trees | High accuracy, handles nonlinearity | Cross-sectional ranking, anomaly detection | Sensitive to tuning |
| Support Vector Machines | Classifier | Effective in high-dimensional space | Stock classification, outperformance tagging | Poor scalability |
| k-Nearest Neighbors (k-NN) | Distance-Based | Simple, non-parametric | Clustering, peer group analysis | Inefficient in large datasets |
| k-Means Clustering | Unsupervised | Regime detection, similarity grouping | Market regime clustering | Requires predefining k |

An expanded classical ML workflow for alpha generation (Fig. 1.) typically encompasses problem definition, data collection, preprocessing, feature engineering, model training, signal generation, portfolio construction, backtesting, risk management, deployment, and feedback loop for model monitoring and updates. These pipelines are designed to capture alpha signals from structured financial features such as momentum, value, volatility, and liquidity.

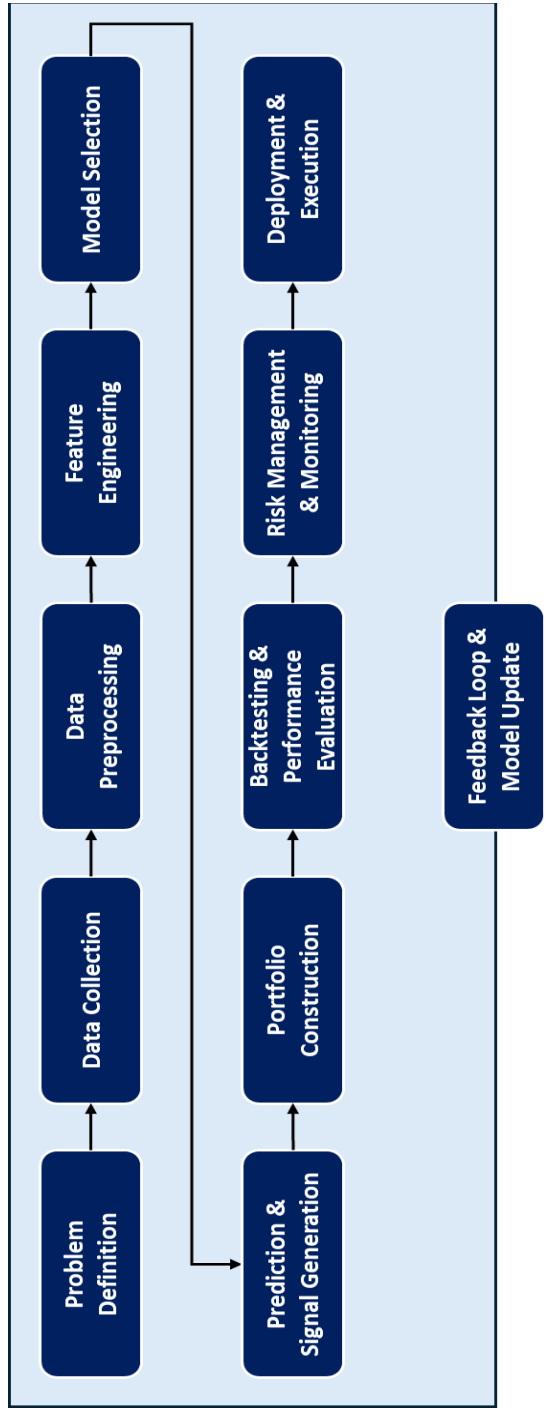

**Fig. 1.** Classical ML Workflow in Alpha Generation

Classical machine learning models offer several advantages in financial modeling, including greater adaptability to complex historical patterns [13,39], improved performance on noisy or nonlinear datasets [14, 29,34], and broad scalability for cross-sectional analysis [35]. However, these benefits come with notable limitations. Most classical approaches rely heavily on manual feature engineering, which introduces domain-specific bias and undermines robustness in evolving market conditions [15, 34].

Additionally, these models often suffer from limited interpretability, particularly when outcomes are driven by ensembles of decision trees or support vectors [39,40]. Their applicability is also largely confined to structured or tabular data formats, making it difficult to incorporate information from natural language or graph-based relationships [26, 41]. These constraints underscored the need for more flexible architectures capable of learning directly from raw inputs, integrating diverse data modalities, and streamlining fragmented model pipelines—requirements that ultimately led to the rise of deep learning.

## 3 Deep Learning for Alpha Generation

Deep learning (DL) introduced a paradigm shift by enabling end-to-end learning from raw inputs and effectively capturing temporal, spatial, and relational patterns through specialized neural network architectures such as Convolutional Neural Networks (CNNs), Recurrent Neural Networks (RNNs), and Graph Neural Networks (GNNs) [27, 42-44]. These capabilities expanded alpha discovery beyond structured signals into complex domains. Despite its strengths, DL also introduces challenges such as overfitting, latency, and opacity [38, 40, 45]. For a summary comparison of DL's evolution within the broader alpha landscape, refer to Stage 4 in the taxonomy (Section 5). The following subsections examine DL's architectural foundations, its progression toward end-to-end and multimodal learning, and the practical and regulatory challenges that affect its deployment in production alpha systems.

### 3.1 Spatio-Temporal Modeling

A key strength of deep learning lies in its ability to capture both temporal dynamics and relational dependencies in financial data—essential for modeling asset co-movements, market regimes, and investor behavior. While the taxonomy in Section 5 outlines the core architectures, Table 3 highlights their practical applications, including Long Short-Term Memory (LSTM) and Gated Recurrent Unit (GRU) networks for time-series forecasting, Convolutional Neural Networks (CNNs) for order book modeling, Graph Neural Networks (GNNs) for capturing cross-asset relationships, and hybrid CNN-GNN models for sector-aware signal fusion.

**Table 3.** Applications of Spatio-Temporal Architectures in Alpha Generation

| Architecture | Use Case | Description |
|---|---|---|
| LSTM / GRU | Time-series forecasting | Models long-term dependencies in noisy price/volume data |
| CNN | Limit order book prediction | Extracts local patterns from structured depth snapshots |
| GNN | Asset relationship modeling | Captures cross-asset correlation and sectoral structure |
| Hybrid (CNN + GNN) | Sector-aware signal fusion | Combines local trends with network-aware features |

These spatio-temporal approaches are increasingly adopted in production trading systems, enabling multiscale modeling across intraday signals, asset networks, and regime shifts. More importantly, they provide the representational foundation for transformer-based and LLM-powered models that fuse language, structure, and time into unified alpha pipelines.

### 3.2 End-to-End and Multimodal Learning in Alpha Generation

Traditional alpha generation pipelines—built as modular sequences of feature engineering, signal modeling, and portfolio optimization—often suffer from integration friction, error propagation, and misalignment between intermediate outputs and final investment objectives [15, 34]. Deep learning addresses these limitations through end-to-end architectures that enable direct optimization from raw or minimally processed inputs to actionable signals [27, 50]. Fig. 2. illustrates a multimodal pipeline for alpha generation, where heterogeneous financial data—such as time-series prices, fundamental indicators, textual sentiment, and asset relationships—are ingested by modality-specific subnetworks and fused into a unified predictive framework [26, 44, 46]. This integrated design not only reduces manual intervention but also enhances adaptability to changing market conditions [25].

**Architectural Shift: From Modular Pipelines to End-to-End Learning**: In conventional quantitative workflows, distinct models are often tasked with preprocessing, feature selection, signal scoring, and risk-adjusted portfolio construction. Each component is tuned in isolation, which can introduce misalignment between intermediate representations and final investment objectives [15, 34]. End-to-end learning frameworks circumvent this issue by jointly training the entire architecture to minimize a target loss function (e.g., cross-sectional return ranking loss or directional accuracy), allowing intermediate representations to evolve dynamically based on downstream task relevance [27, 50]. Such architectures are particularly advantageous in non-stationary environments, where handcrafted features and rigid priors may fail to capture latent dynamics [25, 38]. Models like Temporal Convolutional Networks (TCNs) and deep LSTMs can ingest sequences of historical market data and learn temporal hierarchies without manual lag construction [43, 51]. When coupled with autoencoders or attention-based mechanisms, these systems provide a scalable means of abstracting relevant predictive features across various market regimes [45, 50].

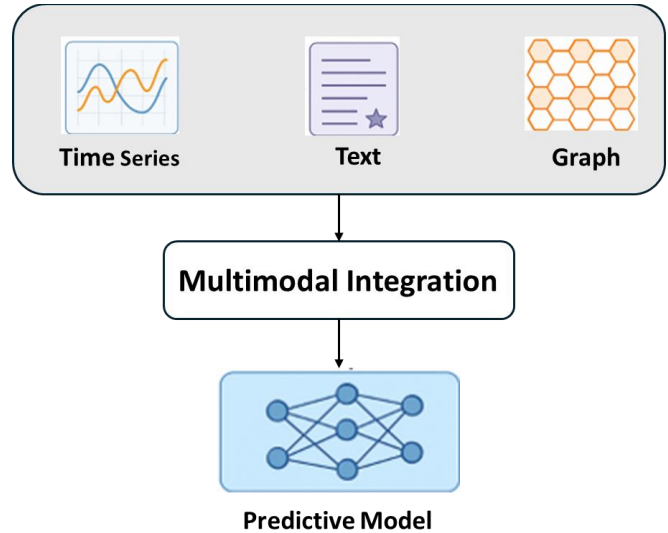

**Fig2.** Multimodal Pipeline for Alpha Generation

**Multimodal Learning for Alpha Generation:** Modern alpha generation increasingly depends on the integration of heterogeneous data modalities—including numerical time-series, textual sentiment from news and earnings calls, company fundamentals, and structural relationships among assets [21, 22, 26]. Multimodal deep learning offers a systematic framework for unifying these diverse inputs within a single predictive model capable of capturing both cross-sectional and cross-modal dependencies [25, 26, 46]. These systems typically employ modality-specific subnetworks—such as Long Short-Term Memory (LSTM) networks for time-series, transformer-based encoders for textual data, and Graph Neural Networks (GNNs) for relational structures—whose outputs are fused through attention mechanisms or transformation layers into a shared latent representation [43, 44, 48]. Table 4 summarizes the roles of these architectures and their contributions to predictive performance, data integration, and interpretability. Representative use cases include combining real-time news sentiment with price signals, integrating linguistic cues from earnings transcripts with structured financials, and aligning macroeconomic events with volatility dynamics [21, 22, 49].

**Table 4.** Deep Learning Modles for Alpha Signal Pipelines [22, 27, 40, 43, 44, 46, 48, 50-52, 54, 55]

| Model / Component | Role in End-to-End System | Input Modality | Use Case in Alpha Generation | Notable Advantage |
|---|---|---|---|---|
| Transformer / BERT-like Encoder | Contextual representation of unstructured text | Text (e.g., news, transcripts) | Sentiment-aware alpha prediction, event-driven trading | Captures long-range language dependencies |
| LSTM (as subnetwork) | Sequential encoder within multimodal pipeline | Time-series | Integrates price/volume trends into fused alpha model | Efficient temporal learning for structured sequences |
| Graph Neural Network (GNN) | Structural encoder for relational data | Asset graphs, supply chains, sector maps | Enhancing classical factors with sector and co-movement context | Models network effects in asset behavior |
| Fusion Layer (e.g., attention, concat) | Modality integration across subnetworks | Combined (text, numeric, graph) | Unified alpha signal generation across diverse modalities | Enables joint reasoning across heterogeneous signals |
| Temporal Fusion Network (TFN) | Cross-modal, time-aware fusion | Asynchronous inputs (macro, policy, price) | Aligns and weights signals with temporal context (e.g., earnings + volatility) | Handles temporal offsets between modalities |
| Autoencoder (AE) | Feature compression, signal abstraction | Structured data (fundamentals, technicals) | Denoising and reducing dimensionality before fusion | Learns compressed latent signals |
| Modality Gating / Attention Maps | Interpretability and dynamic weighting | All (model-internal outputs) | Highlights modality salience under specific market conditions | Supports explainability and adaptive inference |
| SHAP / Attribution Layers | Post hoc interpretability tool | Final alpha predictions | Explains contribution of each modality to predicted alpha | Facilitates auditability and trust in model behavior |

**Architectural Advantages and Limitations:** Multimodal, end-to-end architectures provide several compelling advantages over modular or single-modal frameworks in the context of alpha generation. By jointly optimizing across heterogeneous inputs— such as time-series data, textual sentiment, and relational structures—these models enhance predictive fidelity and reduce reliance on any single signal source [25, 26, 46]. They also offer contextual adaptability, dynamically adjusting the weight of each modality based on prevailing market conditions (e.g., assigning greater importance to text during volatility spikes or crisis events) [21, 49]. However, these benefits come with significant architectural and operational costs. These benefits come with challenges such as synchronization overhead, data leakage risks, and architectural complexity, all of which must be carefully managed in production settings [50, 54]. Furthermore, interpretability remains a major barrier to deployment. Attribution across modalities is often opaque, particularly when fusion occurs in high-dimensional latent spaces. While post hoc tools such as attention heatmaps, modality gating diagnostics, and SHAP-based cross-modal attribution are being explored to improve explainability, they remain in early stages and may fall short of the transparency required for fiduciary-grade alpha systems [40, 52]. As such, the integration of multimodal, end-to-end models into production workflows must balance performance gains with interpretability and governance constraints [30, 45, 53].

**Implications for Production Alpha Systems:** The ability to reason across modalities positions multimodal deep learning as a foundational enabler of next-generation alpha systems. In particular, these models serve as precursors to agentic architectures—where alpha generation is not only data-driven but context-aware, dynamically adjusting strategies in response to market events, narrative shifts, and structural changes [23-25]. As financial firms increasingly seek to operationalize contextual intelligence, natural language understanding, and relational reasoning, multimodal systems represent a critical bridge between classical statistical models and fully autonomous financial agents [18, 19, 26]. Their role will likely expand from isolated signal generators to core engines of adaptive, explainable, and real-time decision support within modern investment platforms.

To illustrate the application of multimodal deep learning in alpha signal generation, this paper presents a custom formulation (Equation 2) where the alpha score is modeled as a nonlinear function of fused embeddings from text, structured signals, and graph-based features:

$$\alpha_i = \sigma(W_t T_i + W_s S_i + W_g G_i + b \;\;) \tag{2}$$

Where $T_i$ represents text embeddings (e.g., from news or earnings calls), $S_i$ denotes structured financial signals (e.g., returns, volatility), and $G_i$ captures graph-based features such as sector correlations. The learnable weights $W_t, W_s, W_g$, along with bias $b$, allow the model to assign relative importance to each modality, while $\sigma$ is a nonlinear activation function (e.g., ReLU or tanh). This composite formulation is informed by multimodal fusion strategies found in recent financial AI research, where structured inputs, text embeddings, and graph-based representations are linearly combined and passed through non-linear activation to yield unified alpha signals [45, 46, 50]. Table 4 summarizes key deep learning components used in end-to-end and multimodal alpha generation pipelines, highlighting their roles, input modalities, and contributions to predictive performance and interpretability.

### 3.3 Practical and Regulatory Challenges in Deploying DL for Alpha Generation

Despite the demonstrated performance of deep learning in financial modeling, its integration into production-level alpha systems remains constrained by several systemic barriers. These include overfitting in non-stationary environments, limited interpretability, latency issues, and organizational conservatism [38, 53, 56]. As emphasized in *Generative AI, Cybersecurity, and Ethics* [30], deploying AI in high-stakes domains like finance requires more than technical accuracy—it demands explainability, resilience, and alignment with institutional governance frameworks. Table 5 summarizes the primary technical and operational challenges currently limiting deep learning's institutional adoption [40, 52, 53], including interpretability tools such as SHAP (SHapley Additive exPlanations), LRP (Layer-wise Relevance Propagation), and Integrated Gradients, as well as regularization techniques like L2 norm (also known as Euclidean norm) penalties.

## 4 The Emergence of Large Language Models

Large Language Models (LLMs) represent a qualitative leap in financial AI—enabling real-time reasoning, tool use, and contextual alpha generation from multimodal data. From FinGPT and BloombergGPT to agentic systems using LangChain and AutoGPT, LLMs now support sentiment extraction, scenario simulation, and task execution [16, 18-20, 23, 24]. Sections 4.1 and 4.2 differentiate between LLMs as predictors and LLMs as autonomous agents. Capabilities such as zero-shot inference, API chaining, and memory-based reasoning mark the transition to true agentic alpha. To contextualize the role of LLMs within the full five-stage evolution of alpha strategies, see Stage 5 in the taxonomy (Section 5). The following subsections examine the application of LLMs as predictive engines for alpha generation, their progression into autonomous, tool-using agents, and a summary of their core strengths, implementation challenges, and operational considerations in production contexts.

### 4.1 LLMs as Alpha Predictors

LLMs handle sentiment analysis, risk tagging, and scenario interpretation across unstructured sources like earnings calls and news [64, 65]. By unifying feature extraction and signal modeling, LLMs reduce engineering overhead and enable faster iteration. Example applications include sentiment scoring, macroeconomic translation, and event detection. While LLMs are increasingly multimodal, architectural details and use cases are covered in section 7.1. Table 6 summarizes core LLM components used in alpha generation systems, highlighting their roles, input modalities, applications, and functional advantages across predictive and agentic workflows.

**Table 5.** Deployment Challenges of Deep Learning in Alpha Generation [52, 53, 55, 57-60]

| Challenge | Description | Implication for Production Use |
|---|---|---|
| Overfitting in Non-Stationary Environments | DL models tend to overfit to noise in environments with structural breaks and low signal-to-noise ratios. Techniques like dropout, early stopping, and L2 regularization offer only partial mitigation | Limits generalization across regimes; increases fragility |
| Interpretability and Regulatory Explainability | Deep architecture is often opaque. Post hoc tools like SHAP (SHapley Additive exPlanations), LRP (Layer-wise Relevance Propagation), and Integrated Gradients offer local insights, but lack causal grounding. | Limits use in audited environments; undermines model trust for fiduciary decisions |
| Latency and Engineering Complexity | Multimodal models require high computational overhead, synchronized pipelines, and often introduce latency unsuitable for high-frequency trading. | Makes real-time deployment infeasible; increases maintenance burden |
| Model Monitoring and Maintenance | Requires mechanisms for concept drift detection, scheduled retraining, and continuous validation pipelines to maintain performance. | Elevates engineering overhead; increases operational risk |
| Organizational Risk Aversion | Firms prioritize interpretable and auditable models aligned with governance policies, often sidelining deep learning despite backtest performance. | DL often confined to research, backtesting, or advisory-only roles |

### 4.2 LLMs as Agents

The transition of LLMs from static predictors to autonomous agents represents a structural shift in financial AI—expanding their role from insight generation to task execution within live decision-making environments [23, 24, 63]. Unlike conventional machine learning models, which typically rely on structured inputs and predefined outputs, agentic LLMs can interpret natural language, plan actions, interact with external tools, and adapt behavior dynamically in response to evolving market conditions [25, 30]. Fig. 3. illustrates a representative workflow of an LLM-based agent in financial contexts, highlighting how these systems transform raw inputs into adaptive decisions through iterative reasoning, memory, and tool integration.

**Table 6.** LLMs Components in Alpha Generation Systems [16, 18, 23-25, 30, 54, 58, 59, 62]

| Model / Component | Role in Alpha Pipeline | Input Modality | Use Case in Alpha Generation | Notable Advantage |
|---|---|---|---|---|
| Generic LLM (e.g., GPT-4) | Foundation model for text understanding and generation | Unstructured text (news, filings, transcripts) | Sentiment extraction, summarization, thematic clustering | Zero/few-shot reasoning; prompt-based flexibility |
| FinGPT / FinBERT | Domain-specific financial text encoder | Financial documents and social media | Earnings call analysis, risk tagging, market tone detection | Finance-trained vocabulary and numeracy |
| LLM Agent Framework (LangChain, AutoGPT) | Task orchestration and tool integration | Natural language, APIs, structured data | Multi-step workflows: portfolio queries, scenario simulation, alerting | Enables dynamic planning, reasoning, and tool use |
| Retrieval-Augmented Generation (RAG) | Augment LLMs with factual grounding | Text + structured knowledge bases | News summarization, compliance search, explainable rationale generation | Reduces hallucination; injects context-specific data |
| LLM-based Copilot (Query Assistant) | Interactive interface for user prompts | Text queries and tabular data | Portfolio analytics, event attribution, decision support | Human-aligned language interface; real-time query handling |
| Prompt Engineering Module | Instruction formatting and context injection | Prompt templates + contextual variables | Improves LLM precision and reproducibility across tasks | Enhances consistency; reduces prompt volatility |
| RLHF / Safety Layer | Output filtering and behavior alignment | Model output layer | Constrains generation to safe, compliant, and relevant responses | Supports auditability and alignment with institutional policy |

Enabled by frameworks such as LangChain, AutoGPT, FinAgent, and BabyAGI, agentic LLMs are capable of:

- Interpreting natural language prompts (e.g., "Summarize today's risks to my portfolio"),
- Planning and sequencing multi-step tasks,
- Calling external APIs (e.g., economic calendars, data feeds),

- Executing SQL queries or portfolio backtests,

- And adapting strategies based on real-time context shifts [23, 24, 63]

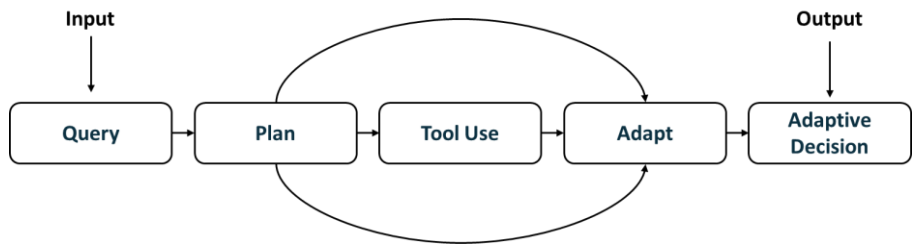

**Fig. 3.** Workflow of an LLM-Based Agent in Financial Contexts

These capabilities allow LLMs to act not merely as forecasters, but as embedded financial co-processors—supporting functions across trading, risk monitoring, compliance, and research. By consolidating perception, reasoning, and execution within a single architecture, agentic LLMs reduce operational fragmentation and augment human decision-making with continuous, contextual intelligence.

**Domain-Specific Intelligence: BloombergGPT and FinGPT:** Generic LLMs often underperform in finance due to vocabulary mismatch, poor numerical reasoning, and limited generalization. Purpose-built models such as BloombergGPT and FinGPT address this gap:

- BloombergGPT is trained on public and proprietary corpora—news, regulatory filings, and terminal data—enhancing performance on financial Q&A, factor tagging, and document summarization [20].

- FinGPT offers a modular, open-source framework, allowing institutions to fine-tune on their own data (e.g., trade logs, chat transcripts), integrate with toolkits, and deploy on-premises under regulatory constraints [18].

These models are now embedded in agents that act as research assistants, risk monitors, and real-time decision-support systems.

**Use Cases in Financial Operations:** LLM agents are rapidly transforming core functions in asset management, hedge funds, and fintechs:

- **Earnings Call Copilots:** Analyze calls in real time, extract sentiment, compare against historical guidance, and recommend directional trades. [21, 22]

- **Event-Driven Trade Agents:** Ingest news and social media feeds to identify market-moving events (e.g., executive resignations), triggering alerts or position adjustments. [47]

- **Query-Driven Research Bots:** Respond to portfolio-level queries (e.g., "Why did volatility spike last week in tech?") by aggregating positions, headlines, and sentiment analysis. [19]

- **Cross-Asset Scenario Simulators:** Evaluate macro "what-if" scenarios (e.g., rate hikes) across equities, FX, and commodities using a blend of textual and quantitative reasoning. [25]

- **Strategy Co-Pilots:** Integrate with portfolio optimizers and backtesting engines to run experiments, fine-tune parameters, and explain changes in factor weights. [54]

These multi-capability agents consolidate workflows that once required siloed systems and analyst teams—compressing cognitive overhead into a single orchestrated interface.

### 4.3    Strengths and Limitations of LLMs in Finance

While LLMs offer transformative potential for alpha generation, their deployment introduces both significant advantages and operational risks. This section synthesizes insights from their roles as predictors (Section 4.1) and agents (Section 4.2), presenting a unified view of their strengths and limitations in institutional finance. Tables 7 and 8 present a balanced overview of the strengths and limitations of large language models (LLMs) in quantitative finance, highlighting their capabilities in multimodal reasoning and rapid prototyping, alongside challenges related to accuracy, scalability, and operational risk.

**Table 7.** Strengths of LLMs in Quantitative Finance [16, 25, 26, 30, 55, 58]

| Strength | Description |
|---|---|
| Multimodal Reasoning | Integrates structured data, time series, and text in a single interaction for holistic insights. |
| Zero/Few-Shot Generalization | Adapts quickly to new asset classes, market regimes, or linguistic shifts. |
| Human-Aligned Explainability | Generates natural language rationales, improving transparency and communication with stakeholders. |
| Rapid Prototyping | Enables prompt-based iteration without retraining, accelerating hypothesis testing. |

To mitigate these risks, production-grade deployments incorporate human-in-the-loop validation, input/output constraints, and continuous monitoring. Techniques such as Retrieval-Augmented Generation (RAG), Reinforcement Learning from Human Feedback (RLHF), and safety-aware agent orchestration further enhance auditability and reliability. LLM agents represent a shift from passive data interpreters to interactive decision engines—capable of analyzing, initiating, and iterating across complex financial workflows. As agent frameworks mature, they are expected to automate research, accelerate scenario planning, and enhance strategic decision-making throughout the investment lifecycle.

**Table 8.** Limitations and Operational Barriers of LLMs [29, 30, 61- 63-65]

| Limitation | Impact |
| --- | --- |
| Hallucination | May produce fluent but false outputs, risking flawed trades or unexplainable decisions. |
| Prompt Instability | Minor prompt changes can yield inconsistent results, reducing reproducibility. |
| Latency and Scalability | Multi-step reasoning and large model size increase inference time and cost, limiting use in HFT. |
| Factual Calibration | Lacks internal confidence estimates; external verification is needed for accuracy. |
| Autonomy Drift | Without constraints, agents may deviate from objectives or loop indefinitely. |
| Security and Compliance | Agentic access to tools or APIs must be sandboxed to prevent misuse or policy violations. |

## 5    5-Stage Taxonomy of Alpha Strategy Evolution

To better understand and benchmark the rapid evolution of alpha generation techniques, this paper proposes a five-stage taxonomy built upon the technical methods discussed in the preceding sections. This taxonomy systematically classifies the progression of investment intelligence—from manually crafted, intuition-driven strategies to fully autonomous, agent-based architectures [2, 25, 30, 63]. Rather than revisiting individual model details, this framework emphasizes the level of automation, modeling complexity, and decision intelligence embedded in each stage. It provides a unified lens through which quantitative researchers, financial institutions, and AI practitioners can assess current capabilities, identify maturity gaps, and prioritize future development pathways. This architecture (Fig.4) presents a unified view of alpha generation evolution, mapping the transition from manual strategies to LLM-based agents through a five-stage maturity taxonomy. It highlights the flow of diverse data inputs, the progression of modeling techniques, and the persistent challenges that shape the end-to-end alpha pipeline [26, 58].

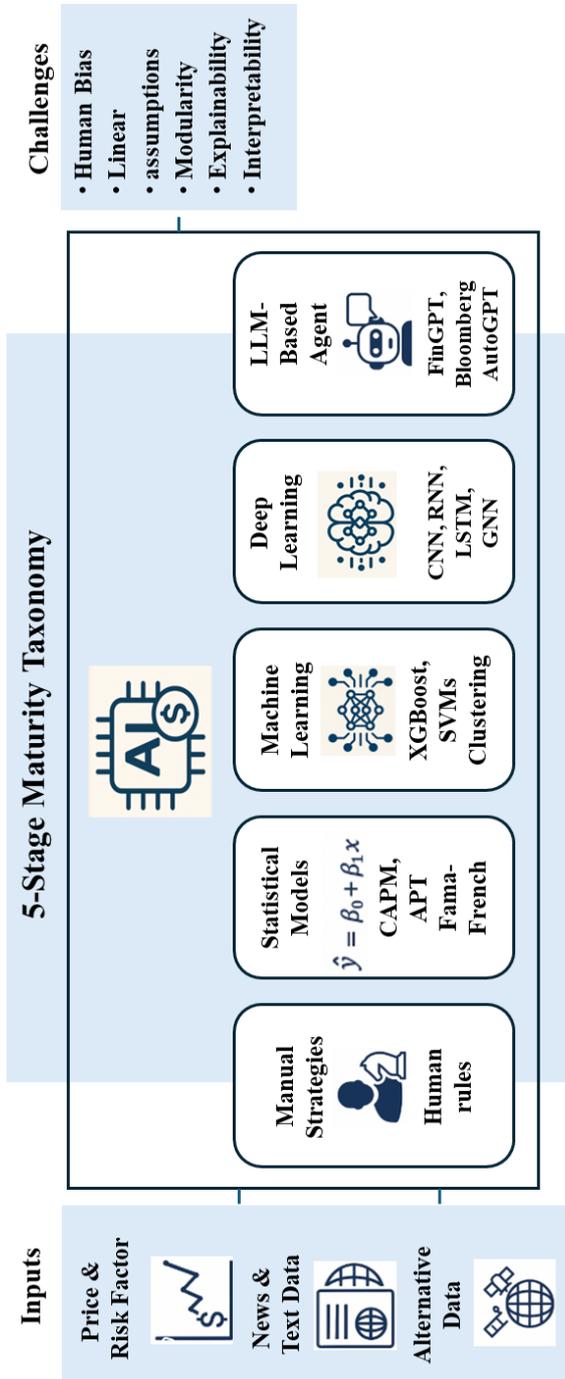

**Fig. 4.** Evolution of Alpha Generation

**Table 9.** Examples of Methodological Shifts in Alpha Generation

| Era | Methodology | Data Type | Key Technologies | Human Involvement | Example Outputs |
|-----|-------------|-----------|------------------|-------------------|-----------------|
| Manual / Fundamental | Heuristics, Intuition | Structured | Analyst Reports | High | Buy/sell recommendations |
| Statistical Modeling | Econometrics, CAPM, APT | Structured | Regression, PCA | Medium | Factor models, risk forecasts |
| Classical ML | Supervised Learning | Structured | Random Forest, SVM | Moderate | Price prediction models |
| Deep Learning | Neural Networks | Semi/Unstructured | CNN, RNN, GNN | Low | Sentiment analysis, trade signals |
| LLM-Driven Agents | Autonomous Agents, Reasoning | Multimodal | GPT-4, FinGPT, Tool-Use APIs | Minimal (Agentic) | Auto-generated strategies |

Inspired by prior taxonomies in AI and fintech innovation studies [66], this structure also serves as a foundation for aligning strategy, governance, and AI infrastructure with the increasing demands of dynamic and data-intensive markets. Table 9 complements this by offering a comparative breakdown of each methodological era in terms of data usage, key technologies, human involvement, and example outputs.

**Stage 1: Manual and Fundamental Alpha (Human-Centric Intelligence):** As discussed in Section 2.1, early alpha strategies relied on human intuition, qualitative reasoning, and discretionary analysis—often informed by company fundamentals, technical indicators, and macroeconomic themes [2-4]. These approaches provided rich contextual insights but lacked scalability and formal testability, marking the starting point of the evolutionary arc.

**Stage 2: Statistical Alpha (Rule-Based Quantification):** Building on these foundations, the transition to statistical modeling introduced formal structure, enabling risk decomposition and systematic factor attribution. Models such as CAPM [1], APT [69], and the Fama-French framework (see Section 2.1) allowed for scalable, backtestable strategies, though they often relied on linearity assumptions and excluded unstructured or real-time data.

**Stage 3: Machine Learning Alpha (Data-Driven Pattern Recognition):** Stage 3 marks the shift to data-driven inference, with classical machine learning (ML) algorithms offering the ability to uncover nonlinear patterns, complex interactions, and latent structures in high-dimensional financial data [13, 14]. Unlike earlier stages, these models derive predictive insights directly from data without predefined factor structures or strong parametric assumptions. Common methods include tree-based ensembles (e.g., Random Forests [34], XGBoost [32]), Support Vector Machines (SVMs) [35], and unsupervised clustering techniques [15]. ML techniques expanded the alpha toolkit to support return ranking, earnings surprise detection, and the early integration of alternative data sources such as sentiment scores and satellite imagery. Key advantages include scalability, adaptability, and the ability to model nonlinearity. However, these models often rely heavily on manual feature engineering and suffer from limited explainability and semantic awareness [40]. While offering statistical power, their opacity and contextual blindness laid the groundwork for the transition to deep learning in Stage 4.

*Stage 4: Deep Learning Alpha (End-to-End Representation Learning):* Stage 4 marks the shift to end-to-end deep learning, where models autonomously learn predictive representations from raw, high-dimensional inputs. As outlined in Section 3, deep neural networks—such as CNNs [48], RNNs, LSTMs [43, 53], and GNNs [44]—enable powerful spatio-temporal and relational modeling. These architectures underpin a broad spectrum of use cases including price forecasting, order book modeling, and asset co-movement inference. A defining capability of this stage is multimodal learning—the fusion of structured data (e.g., prices, fundamentals) with unstructured sources such as earnings transcripts and market news. Rather than reiterating the architectural mechanisms described in Section 4.2, this section emphasizes their strategic impact: improved adaptability to market regimes, richer contextual awareness, and reduced reliance on isolated signals. Despite these advantages, deep learning introduces well-documented challenges—including opacity [52], high data and compute requirements, and model fragility in volatile environments [30]. These limitations, as discussed earlier, catalyze the evolution toward agentic systems capable of incorporating reasoning, memory, and autonomous decision-making, as explored in Stage 5.

**Stage 5: Agentic Alpha (Autonomous Financial Intelligence) :** At the frontier of alpha generation lies the integration of LLMs into autonomous, agentic systems. As discussed in Sections 4, these systems transcend traditional prediction by combining language understanding, tool invocation, and sequential decision-making [25, 63]. Rather than restating prior architectural details, this stage emphasizes the strategic implications of embedding LLM agents into financial workflows. Agentic alpha systems ingest multimodal inputs—including prices, earnings transcripts, macro indicators, and news—and perform tasks such as data retrieval, scenario simulation, and narrative generation [18]. Frameworks like LangChain, AutoGPT [23], FinGPT, and BloombergGPT, previously introduced, enable these agents to plan multi-step actions, integrate APIs, and adapt to real-time market contexts. Core capabilities include tool-augmented autonomy, memory retention, and human-AI collaboration, positioning these agents as cognitive partners rather than static engines. The potential impacts are transformative: real-time signal discovery, dynamic asset reallocation, multilingual analysis, and autonomous portfolio support. However, these systems also introduce new challenges, including hallucination risks [61], latency [64], governance complexity [30], and the need for ethical oversight [59]. Despite these hurdles, agentic alpha represents a convergence of planning, reasoning, and market intelligence—paving the way for adaptive and explainable financial agents built on foundation models.

This taxonomy provides a foundation for evaluating current strategies, identifying technology readiness, and anticipating the next frontier in alpha discovery. It also serves as a framework for aligning AI governance, model risk management, and talent development in quant organizations [30, 66].

## 6    Cross-Cutting Challenges and Meta-Critique

The evolution of alpha generation—from human discretion to statistical modeling, machine learning, and now LLM-driven financial agents—has undeniably accelerated innovation across the investment landscape. Yet this progress also exposes a range of cross-disciplinary challenges that extend beyond individual algorithms. Critical concerns around interpretability, robustness, data integrity, governance, and regulatory compliance increasingly determine whether advanced AI systems can transition from experimental tools to core financial infrastructure. As institutions pursue agentic AI for alpha prediction, strategy execution, and portfolio support, unresolved tensions remain: the trustworthiness of model outputs, adaptability in volatile markets, latent data biases, and the operational readiness of LLMs in real-time, high-stakes environments. As emphasized in *Generative AI, Cybersecurity, and Ethics* [30], addressing these concerns is not solely a technical challenge—it is a prerequisite for deploying safe, transparent, and accountable AI in global financial systems [59,66,68]. The following subsections examine key production challenges in alpha-oriented AI systems, including interpretability, data quality, regulatory alignment, and real-world deployability.

Fig. 5. illustrates a cascading framework of AI risk in financial systems, beginning with fundamental concerns—interpretability, hallucination, latency, and bias—that manifest as downstream effects such as trust degradation and limited auditability. These vulnerabilities heighten exposure to model drift and broader machine learning risks. To counteract these challenges, the diagram emphasizes a multi-pronged mitigation approach: the implementation of explainability methods, structured human oversight, and rigorous robustness testing. Together, these components form the compliance and policy interface required for deploying responsible, transparent, and resilient AI solutions in high-stakes financial contexts.

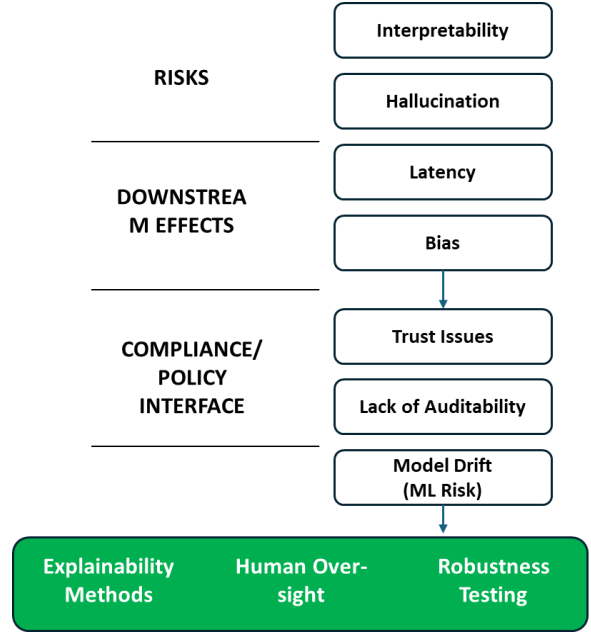

**Fig. 5.** Cross-Cutting Risks and Governance Flow

### 6.1 Interpretability and Trust in AI Models

As alpha generation systems evolve from statistical models to deep learning and agentic large language models (LLMs), interpretability has emerged as a foundational requirement—not only for internal validation but also for regulatory compliance, operational transparency, and stakeholder trust [30]. In high-stakes financial domains, where decisions must be auditable, economically justified, and legally defensible, opaque AI models pose significant barriers to adoption. Building on the limitations of deep learning models discussed in Section 3 and the LLM-specific risks outlined in Section 4—including hallucinations and prompt variability—the broader challenge lies in developing frameworks that quantify trust beyond accuracy metrics. Most explainability tools remain post hoc in nature and approximate influence rather than uncovering a model's true internal logic. This limits their effectiveness during volatile market regimes or compliance audits, where reproducibility and clear rationale are critical [52,53,55].

To address this gap, this paper introduces a composite Trust Score (Equation 3)—a structured metric for evaluating interpretability and reliability in AI-driven alpha systems. The Trust Score aggregates multiple dimensions of explainability and alignment:

$$Trust_i = \omega_1 * Attribution_i + \omega_2 * Stability_i + \omega_3 * (Stability * Factuality_i) + \omega_4 * Alignment_i \qquad (3)$$

Where,

- *Attribution$_i$* measures the consistency and transparency of feature attribution (e.g., via SHAP or attention-based maps),

- *Stability$_i$* evaluates output robustness to input or prompt perturbations,

- *Factuality$_i$* assesses the correctness of outputs relative to ground truth or retrieved knowledge,

- *Alignment$_i$* reflects how well the model's rationale aligns with domain-specific rules, policies, or economic logic.

- The weights $\omega_1$ to $\omega_4$ are tunable, allowing institutions to prioritize dimensions based on governance, compliance, or operational needs.

This score offers a standardized, modular framework to assess model readiness and deployability—complementing predictive performance with a trust-centered evaluation paradigm. As emphasized in *Generative AI, Cybersecurity, and Ethics* [30], the future of AI in finance hinges not just on intelligent models, but on those that can explain and justify their outputs under scrutiny. Embedding trust-aware mechanisms into the design and evaluation of alpha-generating systems will be central to their responsible, scalable deployment in institutional settings.

To quantify model transparency in a structured manner, this paper proposes a SHAP-weighted explainability metric $E$ (Equation 4), derived fro, the cocept of exaplainalble AI (XAI) which aggregates feature-level attributions based on their importance in financial decision-making:

$$E = \sum_{i=1}^{n} |\omega_i * \text{SHAP}_i| \tag{4}$$

Where

- $\omega_i$ denotes the relative importance weight assigned to feature $i$

- *SHAP$_i$* is the SHAP value for feature $i$, reflecting its contribution to the model's prediction,

- $E$ captures the degree to which influential features are also interpretable.

- A higher $E$ indicates that features with greater predictive influence are also more explainable—thereby aligning transparency with decision relevance.

This metric provides a scalable tool for evaluating explainability across models, and can serve as a compliance-aligned diagnostic in institutional finance, particularly where explainability must be audited alongside performance. When integrated into broader trust scoring frameworks, such measures help formalize model governance and bridge the gap between regulatory expectations and AI system design.

## 6.2    Data Availability and Market Adaptivity

The performance and reliability of AI-driven alpha strategies are fundamentally constrained by the nature of the data on which they are trained. Unlike the clean, stationary datasets often assumed in academic machine learning, real-world financial data presents a far more volatile, fragmented, and noisy environment. Market signals are frequently:

- Noisy, reflecting microstructure effects and idiosyncratic fluctuations.

- Non-stationary, due to macroeconomic cycles, regulatory shifts, and changing investor behavior [54, 66].

- Sparse in labels, particularly in emerging markets, niche asset classes, or during rare but high-impact events such as financial crises or geopolitical disruptions [38].

These challenges introduce a persistent disconnect between algorithmic assumptions and market realities. Deep learning models, in particular, often depend on large volumes of homogeneous, stationary data with well-annotated targets—conditions rarely met in financial practice. In addition, global financial data is inherently heterogeneous:

- Disclosure standards vary across jurisdictions, creating asymmetries in data completeness.

- Alternative data sources (e.g., satellite imagery, social media, ESG metrics) differ in structure, granularity, and update frequency [26, 47].

- Event-driven inputs—such as earnings calls or monetary policy announcements—are temporally irregular yet highly impactful.

This fragmentation leads to two systemic risks:

1. Overfitting to outdated historical patterns no longer reflective of current market dynamics;

2. Poor generalization in live deployment, where models must interpret unfamiliar inputs under shifting regimes.

To address these limitations, the field is progressively adopting adaptive learning strategies, including unsupervised and semi-supervised learning, robust backtesting [14], and techniques that explicitly account for concept drift, data leakage, and model fragility [25]. Until AI systems can robustly infer signal amid structural uncertainty and align with live market microstructures, their role in real-time alpha generation will remain limited.

### 6.3   Regulation and Responsible AI in Trading

As AI systems become increasingly integrated into financial workflows—from portfolio optimization to autonomous research agents—regulatory scrutiny is rapidly intensifying. Financial AI is no longer evaluated solely on technical performance; it must also meet standards for ethical alignment, legal compliance, and systemic risk mitigation.

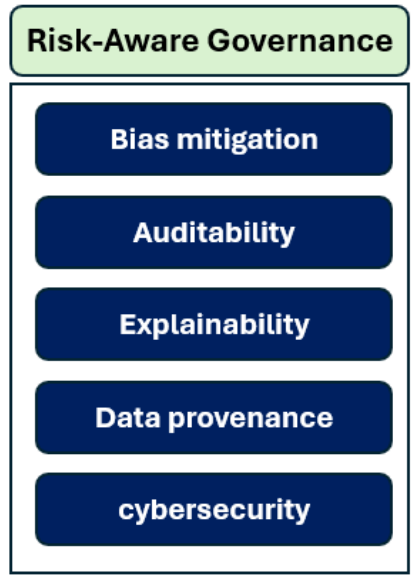

**Fig. 6.** Responsible AI Stack in Trading Systems

Fig. 6 presents a layered view of the *Responsible AI Stack in Trading Systems*, illustrating how risk-aware governance must span the entire lifecycle—from data sourcing and model training to auditability and regulatory oversight. This architecture reflects a shift in both industry and policymaker expectations toward *end-to-end accountability*.

Regulatory frameworks are converging globally on several key themes:

- In the European Union, the proposed AI Act classifies financial services as a *high-risk domain*, requiring documented model logic, risk controls, and human-in-the-loop decision protocols [28].

- In the United States, the SEC and FINRA emphasize algorithmic transparency, particularly for robo-advisors, fraud detection, and automated trading systems [29].

Core pillars of financial AI compliance now include:

- Bias mitigation: Ensuring that models do not systematically disadvantage protected groups in lending, credit, or investment contexts.

- Auditability and explainability: Maintaining the ability to trace outputs back to features, parameters, and data sources [40, 52]

- Data provenance and cybersecurity: Validating the integrity and origin of training data, especially when alternative or third-party sources are involved.

As outlined in *Generative AI, Cybersecurity, and Ethics* [30], responsible AI in finance is not just a technical endeavor—it requires organizational-level controls. These include:

- Human-in-the-loop overrides and escalation pathways [68],

- Fail-safe mechanisms such as kill switches for anomalous model behavior [56]

- Scenario and stress testing to anticipate breakdowns under volatility or edge-case conditions [38].

Systemic risk concerns are also rising. For instance:

- Herding effects may emerge from LLMs trained on similar public data [59]

- Volatility amplification can occur when agentic systems respond reflexively to macroeconomic events [69]

- Cascading failures may result from misinterpretation of ambiguous inputs by autonomous agents operating without sufficient oversight.

To manage these risks, regulators are exploring new mandates for disclosures around model architecture, training data, and degrees of agentic autonomy. Going forward, organizations must adopt AI governance frameworks that not only satisfy regulatory requirements, but also embed the principles of transparency, accountability, and operational resilience across the entire model lifecycle.

### 6.4 Deployment Barriers and Outlook for LLM-Based Alpha Systems

While large language models (LLMs) offer transformative potential in alpha generation, most remain unsuitable for autonomous deployment in high-stakes financial settings. As discussed in Section 4.3, core limitations include hallucination risk, prompt variability, and latency, all of which impact auditability, real-time decision-making, and compliance readiness [61, 62, 64]. Additional deployment constraints include:

- Tool fragility, where multi-step agents may fail due to API dependency or unreliable execution planning [68],

- And resource intensity, with models like GPT-4 incurring high operational and environmental costs that may outweigh their incremental performance gains [64].

Given these challenges, LLMs are currently best positioned as decision-support co-pilots—enhancing tasks such as summarization, risk tagging, and portfolio diagnostics—rather than acting as autonomous trading agents. Looking forward, active research is addressing these gaps through:

- Retrieval-Augmented Generation (RAG) for grounding factual accuracy [25]

- Prompt chaining to reduce brittleness in task execution [63]

- And real-time circuit breakers that flag anomalies before execution [30]

Bridging the deployment gap will ultimately require not just technical refinement, but integration with governance frameworks that embed explainability, human oversight, and operational safeguards—consistent with the broader mandates outlined in Section 4.3 and in *Generative AI, Cybersecurity, and Ethics* [30].

## 7 Future Directions for Research and Practice

The evolution from manual strategies to agentic, AI-powered alpha systems has transformed the landscape of quantitative investing [25, 30, 54]. However, despite impressive breakthroughs, substantial opportunities for innovation remain untapped. Much of today's AI infrastructure in finance remains reactive, brittle, or domain constrained [25, 66, 70]. With continued progress in Large Language Models (LLMs) [16, 18, 20], Reinforcement Learning (RL) [71], AutoML [72], and agent-based simulation, the next generation of alpha systems is expected to be adaptive, self-improving, and contextually aware—capable of collaborating with humans, learning from market interactions, and adjusting dynamically to changing economic regimes [25, 66, 72]. These advances point toward a future where alpha generation is not merely about statistical prediction but about constructing intelligent financial ecosystems—systems that reason, communicate, simulate, and align with both market objectives and regulatory norms [30, 66, 68]. What follows are five critical frontiers that will shape this evolution and provide a roadmap for the research, design, and governance of next-generation financial AI systems. Fig. 7. visualizes this shift by organizing key innovation frontiers into a forward-looking roadmap, including agentic architectures, simulation-based learning, and embedded governance.

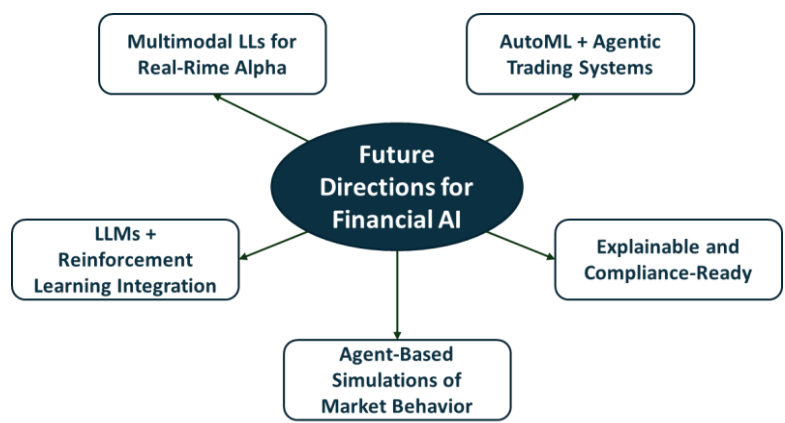

**Fig. 7.** Future Directions

### 7.1 Multimodal LLMs for Real-Time Alpha

Modern alpha systems must integrate diverse financial data—including structured market inputs, time-series patterns, and unstructured text like earnings transcripts. Multimodal Large Language Models (LLMs) provide a unified architecture to fuse these signals using time-series transformers [72], graph-based encoders [44], and document-aware language models [42]. Foundational methods are detailed in Section 4.2. These systems enable context-aware signal generation by dynamically weighting modalities based on relevance (e.g., news salience during crises) [72]. Common applications include:

- Real-time integration of market data and financial news [50]

- Fusion of earnings call sentiment with structured fundamentals [55]

- Reasoning across asset graphs and macro narratives [44]

Key challenges remain—aligning asynchronous data, reducing latency [65], and maintaining explainability [50]. Despite these hurdles, multimodal LLMs offer promising infrastructure for adaptive and robust alpha generation.

### 7.2 Reinforcement Learning for Adaptive Alpha Modeling

Reinforcement Learning (RL) enables agents to improve decision-making policies through repeated interaction with market environments and feedback from realized outcomes. When combined with Large Language Models (LLMs), RL systems can incorporate semantic understanding, scenario simulation, and contextual reward shaping—allowing agents to evolve beyond static rules into adaptive, language-aware strategists [71].

This integration supports the development of AI systems that autonomously learn trading strategies, rebalancing logic, or hedging policies by continually aligning their alpha estimates with observed market signals. A typical update mechanism can be expressed as (Equation 5):

$$\alpha_{t+1} = \alpha_t + \eta(R_t - \alpha_t) \tag{5}$$

where $\alpha_t$ represents the agent's current estimate of alpha, $R_t$ is the realized return or reward, and $\eta$ is a learning rate governing the adjustment magnitude. This formulation captures the essence of **policy refinement via real-time market feedback**, allowing agents to adaptively converge toward high-performing strategies. The update rule, though mathematically similar to temporal-difference learning, is reinterpreted here within the context of alpha generation—treating $\alpha_t$ as a dynamic estimate of strategic performance rather than as a traditional value function.

Despite its promise, deploying RL for alpha generation remains challenging due to instability during training, exploration inefficiencies in sparse-reward settings, and limited interpretability of learned policies. These issues underscore the need for robust governance frameworks and interpretability mechanisms, particularly in production-grade financial systems.

### 7.3 AutoML-Enabled Agentic Trading Systems

Automated machine learning (AutoML) methods can generate, tune, and deploy predictive models with minimal human intervention. Embedding AutoML within agentic LLM systems (e.g., FinAgent, AutoFinAgent) opens the door to self-configuring financial agents capable of autonomously adapting to changing market regimes [72]. These systems can perform hypothesis testing, feature selection, and risk-based optimization without manual orchestration. Nonetheless, they introduce concerns around overfitting, search instability, and the transparency of auto-generated model logic.

### 7.4 Agent-Based Simulations of Market Behavior

Agent-based modeling (ABM) offers a testbed for understanding emergent behaviors in financial ecosystems. LLM-powered agents can be instantiated with heterogeneous objectives and reasoning capabilities to simulate real-world dynamics such as market shocks, regulatory changes, or behavioral herding [68, 71]. These simulations can support systemic risk analysis and regulatory stress testing. The key obstacles include ensuring economic realism, calibrating agents to empirical distributions, and scaling simulations to institutional complexity.

### 7.5 Explainable and Compliance-Ready AI

As financial AI becomes increasingly autonomous, its outputs must remain explainable, auditable, and aligned with fiduciary standards. Techniques such as SHAP, LIME, and counterfactual reasoning help translate complex model behavior into human-interpretable terms [45, 55]. Domain-specific rationalization using natural language generation further enhances transparency and stakeholder trust. However, compliance-ready deployment requires not only model-level explainability but also robust infrastructure for data provenance, reproducibility, and governance [10].

Together, these directions chart a cohesive roadmap for the next generation of alpha generation systems—moving from static, model-centric pipelines toward adaptive, context-aware, and ethically grounded AI architectures. Multimodal LLMs offer the perceptual breadth to integrate diverse financial signals, while reinforcement learning introduces strategic depth through iterative feedback. AutoML frameworks promise scalability and self-optimization, and agent-based simulations open new frontiers for understanding systemic risk and emergent behavior. Crucially, the viability of these innovations will hinge on explainability and regulatory alignment. Future research must therefore balance innovation with interpretability, ensuring that advanced financial AI remains not only powerful but also trustworthy and accountable within real-world market ecosystems.

# 8 Conclusion

This paper has traced the evolution of alpha generation from human-driven strategies to intelligent, agentic systems powered by large language models. By introducing a structured five-stage taxonomy—spanning manual heuristics, statistical modeling, classical machine learning, deep learning, and autonomous financial agents—a unified framework is presented to explain how predictive intelligence in finance has matured across modeling complexity, automation, and contextual adaptability. Beyond technical advancements, the paper identifies key cross-cutting challenges—such as interpretability, market adaptivity, data fragility, and regulatory alignment—that continue to constrain deployment in high-stakes financial environments. To address these gaps, trust-aware evaluation metrics are proposed, including the composite Trust Score and SHAP-weighted explainability measures, to support model governance and compliance readiness. Looking ahead, the future of alpha generation lies in architecting systems that are not only accurate but also adaptive, explainable, and operationally resilient. Innovations in multimodal LLMs, reinforcement learning, AutoML, and agent-based simulation will be central to this transformation. However, real-world viability will depend on embedding governance, transparency, and ethical safeguards into the core of financial AI infrastructure. The proposed taxonomy serves not only as a historical roadmap but also as a foundation for designing next-generation alpha systems that are intelligent, accountable, and institutionally scalable.

## Declarations

**Availability of data and materials**

Not applicable

**Competing interests**

The author declares that he has no competing interests.

**Funding**

Self-funded.

**Authors' contributions**

The author read and approved the final manuscript.

**Acknowledgements**

Not Applicable.

**Author Information**

Dr. Mohammad Rubyet Islam is the author of *Generative AI, Cybersecurity, and Ethics* (Wiley, 2024. With a career spanning three continents and multiple industries, Dr. Islam is a globally recognized leader in Artificial Intelligence and Machine Learning, bridging the gap between cutting-edge academic research and large-scale enterprise AI applications. He has directed AI/ML strategies for numerous Fortune 100 and Global 500 companies, and has led transformative projects in finance, government, healthcare, defense, and retail. He has spearheaded AI/ML initiatives across government (e.g., U.S. Department of Commerce, US Airforce), finance (e.g., Capital One, Fannie Mae), and leading consulting firms (e.g., Booz Allen Hamilton, Deloitte, Lockheed Martin, Raytheon). In the financial domain, his work has focused on developing advanced ML solutions for alpha generation, fraud detection, wealth management, credit risk modeling, and stock prediction. Dr. Islam holds a Ph.D. in Machine Learning and Prognostics & Health Management from the University of Maryland, College Park, MD, USA, in addition to advanced degrees from institutions in Canada, the UK, and Bangladesh. He currently serves as a Generative AI Solutions Architect for the U.S. Government and as an adjunct professor at George Mason University, where he teaches Applied Machine Learning and Natural Language Processing. He has also taught at the University of Maryland, College Park. His scholarly contributions include publications in leading peer-reviewed research journals and conferences. He has served as a reviewer for RSS, ICLR, Wiley, and for journals published by Elsevier and Springer. For his contributions to technology, research, and education, he was honored in Marquis Who's Who in America (2024–2025). To learn more, visit his professional bio: https://ray-islam.github.io/